\documentclass[conference]{IEEEtran}
\IEEEoverridecommandlockouts
\usepackage{cite}
\usepackage[T1]{fontenc}

\usepackage{graphicx}
\usepackage{subcaption}
\usepackage{multirow}
\usepackage{placeins}
\usepackage{tikz}
\usetikzlibrary{positioning,fit,arrows.meta,shapes.multipart}
\usepackage[numbers,sort&compress]{natbib}
\usepackage[hidelinks]{hyperref}
\usepackage{booktabs}
\usepackage{siunitx}
\usepackage[font=small,labelfont=bf]{caption}

\usepackage{amsmath,amssymb,amsfonts}
\usepackage{algorithm}
\usepackage{algpseudocode}
\usepackage{graphicx}
\usepackage{textcomp}
\usepackage{xcolor}
\def\BibTeX{{\rm B\kern-.05em{\sc i\kern-.025em b}\kern-.08em
    T\kern-.1667em\lower.7ex\hbox{E}\kern-.125emX}}
\begin{document}

\title{When Privacy Isn’t Synthetic: Hidden Data Leakage in Generative AI Models}

\author{
\IEEEauthorblockN{S~M~Mustaquim}
\IEEEauthorblockA{
\textit{Department of Mathematical Sciences} \\
\textit{University of Texas at El Paso} \\
El Paso, TX, USA \\
smustaquim@miners.utep.edu}
\and
\IEEEauthorblockN{Anantaa~Kotal}
\IEEEauthorblockA{
\textit{Department of Computer Science} \\
\textit{University of Texas at El Paso} \\
El Paso, TX, USA \\
akotal@utep.edu}
\and
\IEEEauthorblockN{Paul~H.~Yi}
\IEEEauthorblockA{
\textit{Department of Radiology} \\
\textit{St. Jude Children’s Research Hospital} \\
Memphis, TN, USA \\
paul.yi@stjude.org}
}
\maketitle

\begin{abstract}
Generative models are increasingly used to produce privacy-preserving synthetic data as a safe alternative to sharing sensitive training datasets. However, we demonstrate that such synthetic releases can still leak information about the underlying training samples through structural overlap in the data manifold. We propose a \textbf{black-box membership inference attack} that exploits this vulnerability without requiring access to model internals or real data. The attacker repeatedly queries the generative model to obtain large numbers of synthetic samples, performs unsupervised clustering to identify dense regions of the synthetic distribution, and then analyzes cluster medoids and neighborhoods that correspond to high-density regions in the original training data. These neighborhoods act as proxies for training samples, enabling the adversary to infer membership or reconstruct approximate records. Our experiments across healthcare, finance, and other sensitive domains show that cluster overlap between real and synthetic data leads to measurable membership leakage—even when the generator is trained with differential privacy or other noise mechanisms. The results highlight an under-explored attack surface in synthetic data generation pipelines and call for stronger privacy guarantees that account for \textit{distributional neighborhood inference} rather than sample-level memorization alone, underscoring its role in privacy-preserving data publishing.  Implementation and evaluation code are publicly available at \href{https://github.com/AnantaaKotal/Cluster-Medoid-Leakage-Attack}{\texttt{github.com/Cluster-Medoid-Leakage-Attack}}.

\end{abstract}


\section{Introduction}

Data drives scientific progress and technological innovation, but sharing it safely remains a major challenge. History has shown that attempts to anonymize or ``sanitize'' datasets often fail once those datasets are released. In 2006, AOL published what it believed to be an anonymized log of search queries; within days, journalists linked many of those records to real individuals \cite{barbaro2006aol}. In 2007, researchers demonstrated that Netflix’s ``anonymized'' movie ratings dataset could be cross-referenced with public data to uncover personal viewing histories \cite{narayanan2008robust, wired2007netflix}. More recently, Strava’s global ``heat map'' of fitness tracker data inadvertently revealed the locations of U.S. military bases when soldiers’ jogging routes appeared clearly in otherwise remote regions \cite{hern2018strava}.   These examples highlight a broader pattern: privacy protections that appear sound in theory can fail in practice. Traditional anonymization techniques can be undone when datasets are combined with external information. As data becomes richer and analysis tools grow more powerful, the risk of re-identification increases. Once data is shared, even in seemingly safe form, it can interact with other sources in unpredictable and revealing ways.  

In response to these challenges, \textit{synthetic data} has gained traction as a potential solution. The idea is simple: instead of releasing real records, organizations train a generative model on sensitive datasets and then release artificial records sampled from that model. These synthetic samples are designed to preserve the statistical properties of the original data—so that analyses remain valid—while breaking any direct link to real individuals. In theory, this approach separates data utility from privacy risk.  Synthetic data has clear appeal. In \textbf{healthcare}, it promises to enable research on patient outcomes without exposing personal health information \cite{goncalves2020generation}. In \textbf{finance}, it allows banks to collaborate on fraud detection or risk modeling without revealing confidential customer transactions \cite{patki2016sdv}. Government agencies have also explored this idea—for example, the U.S. Census Bureau has investigated the use of synthetic microdata to balance transparency with privacy \cite{hawes2020census}. Commercial interest has followed suit: startups such as Mostly AI, Hazy, and Gretel have raised significant investment to make synthetic data the foundation of privacy-preserving analytics \cite{forbes2021synthetic}.  

This narrative—that synthetic data can be both realistic and private—is compelling. It promises a future where organizations can innovate with data while protecting individual privacy. But an important question remains: \textbf{how private is synthetic data, really?}  

Recent research has shown that generative models, even when designed to produce novel samples, are not immune to privacy risks. Some models \textit{memorize} examples from their training data and can accidentally reproduce them. Membership inference and model inversion attacks have also shown that adversaries can infer whether specific records were used for training or reconstruct sensitive attributes. These attacks suggest that synthetic data does not guarantee privacy by default.  

However, most existing attacks assume a powerful adversary, one with access to the model’s parameters, architecture, or partial training data. In many real-world deployments, such assumptions do not hold. Consider a black-box setting where an attacker has no knowledge of the model internals or training records and no auxiliary background information. The only capability is to repeatedly query the model and observe the synthetic samples it produces. At first glance, this setting appears safe, if all the data released is artificial, what could possibly be inferred about real individuals?  

In this paper, we show that this intuition is misleading. Even in a strictly black-box setting, synthetic data can leak information about the original training set through \textit{distributional overlap}. Figure~\ref{fig:attack-diagram} illustrates the key idea: clusters in the synthetic data tend to align with high-density regions of the real training data. By repeatedly sampling from the generator, an attacker can uncover this structure, identify cluster centers (medoids), and exploit the dense neighborhoods around them to infer or approximate real training samples. These clusters act as \textit{signatures} of the underlying data distribution, exposing traces of the original dataset despite the synthetic outputs.  

\begin{figure}[t]
    \centering
    \includegraphics[width=\linewidth]{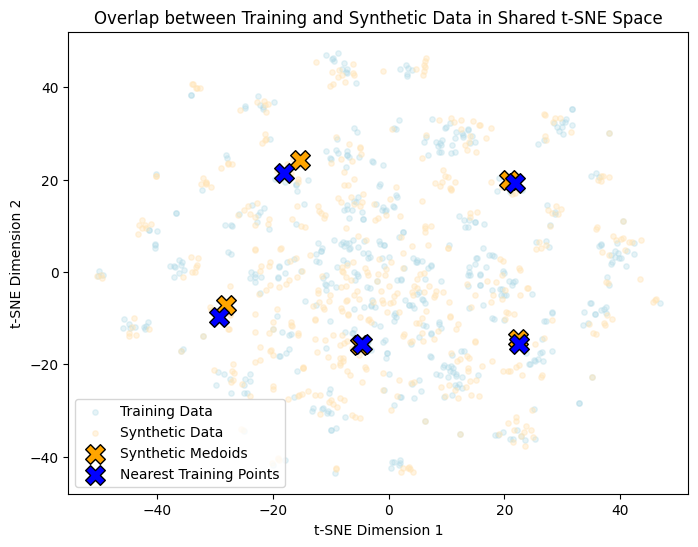}
    \caption{
        Visualization of how synthetic data can unintentionally reproduce the structure of its training data. Even though the generative model produces “new” records, their positions reveal that the generator has learned specific structures from the original dataset. These structural echoes enable an attacker to infer information about the training data without ever accessing it directly.
    }
    \label{fig:attack-diagram}
\end{figure}

A crucial observation is that this weakness is \textbf{model-agnostic}. We observe similar leakage across a wide range of deep generative architectures—GANs, VAEs, diffusion models, and large language models (LLMs). The problem does not stem from direct memorization but from how these models learn the data manifold itself: to generate realistic samples, they must approximate the true data distribution closely enough that their cluster structure mirrors the original. This alignment, while beneficial for fidelity, introduces a new avenue for privacy leakage.  

\textbf{Our contributions are threefold:}  
\begin{enumerate}
    \item We introduce a formal \textbf{black-box threat model} for synthetic data privacy, where the attacker has no access to model internals, training data, or side information.  
    \item We present a \textbf{clustering-based inference attack} that leverages the geometric structure of synthetic data to identify neighborhoods correlated with real training points.  
    \item We evaluate our attack across \textbf{healthcare, finance, and other sensitive domains}, demonstrating that even ``privacy-preserving'' synthetic data can reveal membership information and reconstruct sensitive attributes under strong black-box assumptions.  
\end{enumerate}  

Together, these findings expose an underexplored vulnerability in synthetic data generation. While synthetic data is often marketed as a privacy solution, our results show that the boundary between ``synthetic'' and ``real'' is blurrier than it seems. Distributional overlap between synthetic and training data can create \textit{silent leaks}—subtle but significant exposures of private information. These results call for a rethinking of privacy guarantees in generative models and for new defense strategies that go beyond memorization to address \textbf{neighborhood-level leakage}.

 Implementation and evaluation code are publicly available at \href{https://github.com/AnantaaKotal/Cluster-Medoid-Leakage-Attack}{\texttt{github.com/Cluster-Medoid-Leakage-Attack}}.

\section{Related Work}

\subsection{Privacy, Anonymization, and Their Limits}

Classical anonymization techniques remove direct identifiers or enforce syntactic protections such as $k$-anonymity, $\ell$-diversity, and $t$-closeness \cite{sweeney2002kanon, machanavajjhala2007ldiv, li2007tclose}. While historically influential, these approaches are fragile against linkage with auxiliary information and against high-dimensional sparsity \cite{ narayanan2008robust}. Re-identification case studies have repeatedly shown that even ``sanitized'' datasets can be triangulated to reveal individual identities \cite{narayanan2008robust}.  

Differential privacy (DP) introduced a formal, attack-agnostic framework that bounds the influence of any single record on a released output \cite{dwork2006calibrating}. DP has seen real-world deployments, notably in the 2020 U.S.\ Census \cite{ hawes2020census}, but practical trade-offs remain: utility degradation, cumulative privacy loss, and complex accounting \cite{kifer2012rigor, bun2016concentrated}. These limitations have motivated the turn toward \textit{synthetic data}, which aims to preserve distributional structure while reducing direct identity risk \cite{raghunathan2003mice}.

\subsection{Synthetic Data Generation}

Tabular synthesis methods can be grouped into \textit{parametric} approaches—explicitly modeling $p(\mathbf{x})$—and \textit{non-parametric or neural} approaches that learn high-capacity sampling procedures.

Parametric methods include Bayesian networks, copulas, and model-based imputations. PrivBayes learns a DP-protected Bayesian network and samples from privatized conditional probability tables \cite{zhang2014privbayes}. Copula-based models decouple marginals from dependence, using Gaussian or vine copulas to capture correlations \cite{nelsen2006copulas, vines2002copula}. Such approaches, implemented in frameworks like SDV’s \textit{GaussianCopula} and \textit{CopulaGAN}, balance interpretability with limited flexibility on complex, high-order interactions \cite{patki2016sdv}.  

Deep generative models dispense with explicit distributional forms. Adversarial methods such as TableGAN and CTGAN address mixed types and rare categories in tabular data \cite{park2018tablegan, kotal2024kinetgan, kotal2022privetab}. Variational approaches (TVAE) use latent variable models with continuous relaxations for categorical fields \cite{xu2019ctgan}. Diffusion-based synthesizers like TabDDPM adapt denoising score matching to heterogeneous tables, improving minority-mode coverage \cite{kotelnikov2022tabddpm}. For temporal data, TimeGAN enforces temporal coherence across sequences \cite{yoon2019timegan}.  

Recently, large language model (LLM)-style generators treat tables as token sequences. GReaT tokenizes mixed-type attributes for conditional or unconditional row generation, while RealTabFormer applies transformer decoders with schema-aware constraints \cite{borisov2023great, realtabformer2023}. These models offer controllability via prompts and schema rules.  

Across all families, practical synthesizers rely on mixed-type encodings, conditional sampling for constraints, and post-processing to enforce validity. Tooling such as SDV and SDMetrics provides standardized fidelity, coverage, and privacy-risk metrics \cite{patki2016sdv, SDMetrics2020}. In our experiments, we evaluate representative generators—PrivBayes (DP), PATE-GAN (DP-GAN) \cite{jordon2018pategan}, CTGAN/TVAE (GAN/VAE), TabDDPM (diffusion), and LLM-based GReaT—to analyze how modeling assumptions affect privacy leakage under black-box access.

\subsection{Privacy Attacks on Machine Learning}

A central question in privacy research is whether models leak information about their training data. \textit{Membership inference} (MI) attacks test whether a sample was used in training by exploiting output confidence or generalization gaps \cite{shokri2017membership, yeom2018privacy, kotal2023privacy}. Such attacks work in both white- and black-box settings and are amplified by overfitting or calibration errors \cite{sablayrolles2019whitebox}. \textit{Attribute inference} predicts hidden features from model outputs \cite{melis2019exploiting, kotal2023privacy}, while \textit{model inversion} reconstructs representative or input-consistent records \cite{yang2019inversion}. Data-extraction attacks further show that large models can regurgitate rare training sequences verbatim \cite{carlini2019secret}.  

Generative models add further privacy challenges. Hayes et al.\ adapted MI to GANs via the LOGAN framework \cite{hayes2019logan}, and Hilprecht et al.\ used shadow modeling to evaluate MI against VAEs and GANs \cite{hilprecht2019monte}. Subsequent studies explored how sampling temperature, diversity, or truncation affect leakage \cite{chen2020ganleaks, liu2020stochastic}. Recent audits of diffusion and text models document reproduction of rare or even exact training samples \cite{somepalli2022diffusionregurg}. For tabular generators, MI and attribute leakage have been observed, particularly for minority strata and rare attribute combinations \cite{stadler2022synthetic, jordon2018pategan, kotal2024differentially}. Overall, these studies show that generative models can encode training data in subtle ways, depending on data diversity, regularization, and sampling.  

\textbf{Position of this work:}   Our study builds on these foundations but targets a stricter adversarial regime: a \textit{black-box} attacker with no visibility into model parameters, training data, or side information—only the ability to sample synthetic outputs. Prior MI and inversion work typically assumes white- or gray-box access or auxiliary datasets. We instead ask whether the \textit{geometric structure} of synthetic data alone can leak private information. By clustering synthetic samples, we show that even minimal black-box access reveals latent overlap between synthetic and real distributions. Our framework is model-agnostic and complements extraction or model-stealing attacks: rather than replicating the generator, we infer properties of the hidden training distribution itself. While formal differential privacy can mitigate such risks, our findings reveal that without it, even a lightweight clustering analysis can expose sensitive information through structural patterns in synthetic data.

\section{Methodology}
\label{sec:methodology}

\subsection{Overview}

Figure~\ref{fig:attack-diagram} illustrates the high-level intuition behind our proposed \textbf{black-box cluster-overlap attack} on privacy-preserving synthetic data generators.  
The attack is motivated by a simple but powerful observation: when a generator learns from real data, it often retains the overall structure of the training distribution.  
Even if no individual record is memorized, the synthetic samples it produces may form clusters that align closely with real data clusters.  
By repeatedly sampling from the generator and analyzing the geometry of these samples, an attacker can detect such overlap—revealing subtle forms of memorization and membership leakage without ever accessing the training data or model internals.

\begin{figure*}[t]
    \centering
    \includegraphics[width=\textwidth]{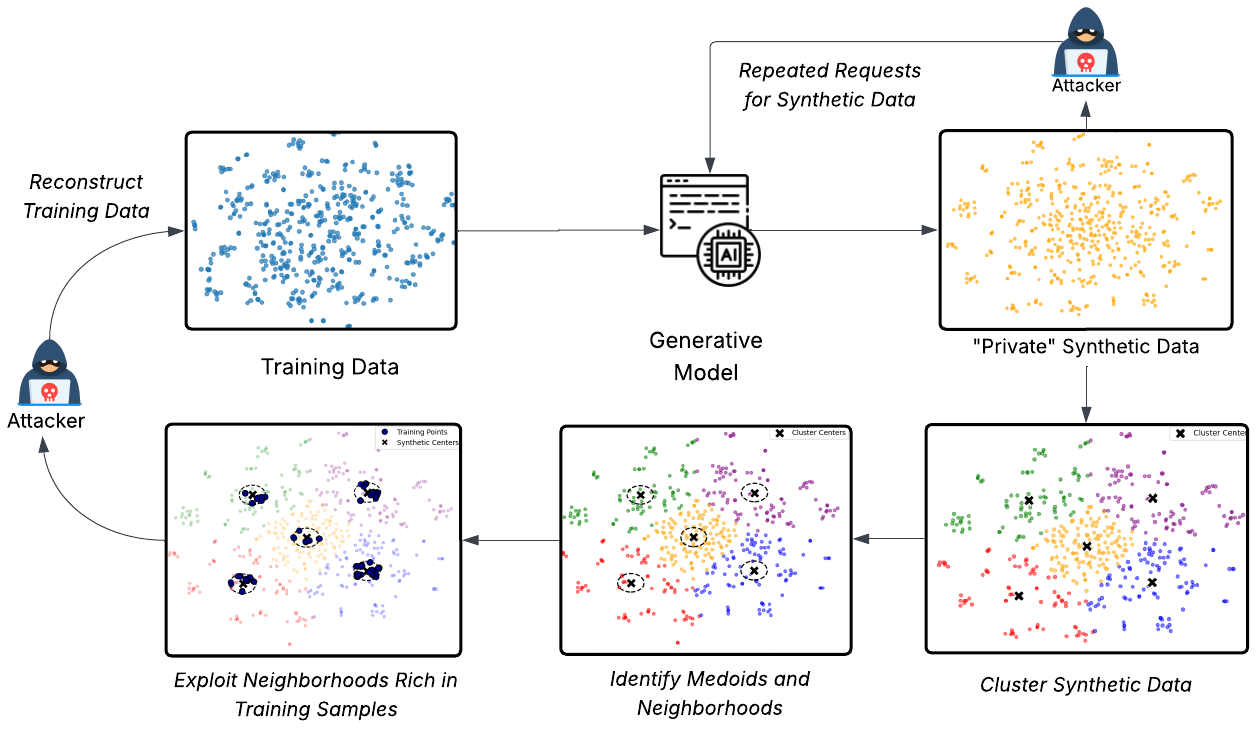}
    \caption{\textbf{Conceptual overview of the black-box cluster-overlap attack.}
    The attacker repeatedly queries the synthetic data generator to collect synthetic samples, clusters them to uncover dense regions, and identifies cluster centers (medoids). 
    Overlaps between synthetic and real data clusters indicate memorization or membership leakage, even when the attacker never sees the real training data.}
    \label{fig:attack-diagram}
\end{figure*}

\subsection{Problem Setting}

Let $\mathcal{D}_{\mathrm{real}}$ denote a sensitive dataset (e.g., patient health records or financial transactions) drawn from an unknown distribution $P$.  
A data owner trains a generative model $G$ to learn $P$ and then releases $G$—or its synthetic samples—for downstream use.  
The adversary’s objective is to determine whether any information about $\mathcal{D}_{\mathrm{real}}$ can be inferred from black-box access to $G$.  
In this setting, the adversary:
\begin{itemize}
    \item has no access to $\mathcal{D}_{\mathrm{real}}$ or $P$,
    \item cannot view $G$'s parameters, architecture, or gradients, and
    \item may only query $G$ to obtain synthetic samples.
\end{itemize}

This reflects a realistic deployment scenario in which organizations publish synthetic data through APIs or share pre-trained generators.  
Although such models are often labeled as ``privacy-preserving,'' they may still expose statistical fingerprints of their training data.  

Formally, the attacker can draw an arbitrary number of synthetic samples
\[
\mathcal{S}_G = \{ s_1, s_2, \dots, s_M \}, \quad s_i \sim G(\cdot),
\]
and analyze their structure.  
If certain regions in the synthetic space are overrepresented or highly compact, they may correspond to dense regions in the original dataset, implying leakage through \emph{distributional overlap}.  
Our goal is to identify and quantify this overlap using only $\mathcal{S}_G$.

\subsection{Attack Intuition}

At its core, the attack treats the synthetic data distribution as a distorted mirror of the real one.  
If the generator memorizes local details, then its synthetic samples will form tight clusters that align with neighborhoods of the training data.  
The attacker can exploit this by reconstructing the approximate ``shape'' of the training distribution from synthetic samples alone.

The attack proceeds by repeatedly querying the generator, collecting synthetic data, identifying clusters of high density, and extracting representative synthetic points—called \textit{medoids}.  
These medoids summarize the dominant regions of the generator’s learned manifold.  
When compared (offline) to the real data, they reveal whether the synthetic clusters are unnaturally close to true samples, indicating possible memorization.

This process yields a set of interpretable and quantitative metrics describing how much of the training distribution is implicitly exposed through synthetic outputs.  
In practice, these analyses can be performed entirely with off-the-shelf clustering tools, requiring no model internals and no adversarial training.

\subsection{Attack Framework}

The proposed \textbf{Cluster–Medoid Leakage Attack (CMLA)} operates in five main stages, each corresponding to a capability that is realistically available to an external adversary.  

\paragraph{Sampling synthetic data}
The attacker issues repeated queries to $G$ to collect a large corpus of synthetic samples, denoted $\mathcal{S}_G$.  
In most synthetic data APIs, sampling is unconstrained—users can request thousands or millions of samples.  
This allows the attacker to approximate the model’s learned distribution with high fidelity.

\paragraph{Creating a shared representation}
Because tabular datasets often contain mixed numeric and categorical attributes, the attacker encodes each record into a shared feature space using a deterministic function $\Psi(\cdot)$.  
Categorical variables are one-hot encoded, continuous features are standardized, and all dimensions are rescaled to comparable magnitudes.  
The resulting vectors lie in a common metric space with a fixed distance measure $\delta$ (e.g., Euclidean or Gower distance).  
This ensures consistent comparisons between synthetic points and, later, between synthetic and real points.  

To improve stability, a dimensionality reduction step (e.g., PCA, UMAP, or t-SNE) can be applied to make geometric structure more visible and reduce noise.  
This embedding is learned solely from synthetic samples, so it remains consistent with the black-box assumption.

\paragraph{Clustering synthetic samples}
Next, the attacker applies unsupervised clustering (e.g., DBSCAN or HDBSCAN) to the encoded synthetic samples.  
Each cluster corresponds to a region where the generator repeatedly emits similar records, representing a \textit{mode} of the synthetic distribution.  
Noise points—synthetic records that occur rarely or inconsistently—are discarded, since they add little information about memorization.  
The number of clusters $K$ is determined automatically by the clustering algorithm and depends on the complexity of the generator’s output distribution.

\paragraph{Extracting medoids}
Within each cluster, the attacker identifies a \textbf{medoid}—the synthetic record most centrally located within that cluster.  
Unlike centroids, which may be non-existent averages, medoids correspond to actual synthetic samples, making them realistic representatives of the generator’s behavior.  
Each medoid can be viewed as a “snapshot” of a synthetic mode: if that mode corresponds to a true region in the training data, the medoid will lie close to one or more real records.  
The resulting set of medoids $\mathcal{M} = \{ m_1, \dots, m_K \}$ acts as a compressed fingerprint of the generator, capturing where it places density mass.

\paragraph{5. Evaluating proximity and overlap.}
Finally, to evaluate potential leakage, the attacker (or auditor) computes how close each medoid lies to real data points.  
Although a real attacker would not have access to $\mathcal{D}_{\mathrm{real}}$, this comparison is useful for empirical auditing.  
The key quantity is the \textbf{nearest-neighbor distance} from each medoid to the closest real record:
\[
d_{\min}(m_i) = \min_{x \in \mathcal{D}_{\mathrm{real}}} \delta(\Psi(m_i), \Psi(x)).
\]
A small $d_{\min}(m_i)$ indicates that a synthetic mode lies within a tight neighborhood of the real data—precisely the signature of localized memorization.

\subsection{Attack Success Rate and Coverage Metrics}

Because raw distances are not easily interpretable across datasets, we summarize them using thresholded metrics that describe how much of the synthetic distribution intrudes into the real one.

\paragraph{Attack Success Rate (ASR).}
We define the \textbf{Attack Success Rate} at distance threshold $\tau$ as the proportion of medoids that lie within radius $\tau$ of any real record:
\[
\mathrm{ASR}(\tau) = \frac{1}{K} \sum_{i=1}^{K} \mathbf{1}\!\left[d_{\min}(m_i) < \tau\right].
\]
Intuitively, $\mathrm{ASR}(\tau)$ answers the question: “What fraction of the generator’s mode representatives fall suspiciously close to the training data?”  
Plotting $\mathrm{ASR}(\tau)$ as a function of $\tau$ yields an interpretable risk curve.  
Sharp early rises (high ASR at small $\tau$) signal significant overlap, suggesting that the generator reproduces regions of the training distribution with high fidelity—potentially compromising privacy.

\paragraph{Coverage.}
Complementary to ASR, we define \textbf{coverage} as the proportion of real records that lie within a given radius of at least one synthetic medoid:
\[
\mathrm{Cov}(\tau) = \frac{1}{|\mathcal{D}_{\mathrm{real}}|} \sum_{x \in \mathcal{D}_{\mathrm{real}}}
\mathbf{1}\!\left[ \min_{m_i \in \mathcal{M}} \delta(\Psi(x), \Psi(m_i)) < \tau \right].
\]
Coverage measures how much of the real data space is “shadowed” by synthetic clusters.  
High coverage at small $\tau$ implies that synthetic modes densely populate the same neighborhoods as real data, signaling potential leakage.

\paragraph{Interpretation.}
Together, ASR and coverage provide complementary perspectives.  
ASR captures the aggressiveness of the synthetic distribution—how many of its modes encroach upon real data—while coverage captures the vulnerability of the real data—how much of it lies under synthetic influence.  
Both are monotonic with respect to $\tau$ and can be used to compare privacy risks across different generators or training configurations.  

These metrics translate the geometric concept of overlap into quantifiable, dataset-agnostic measures of leakage.  
They also allow standardized reporting: privacy auditors can fix $\tau$ values corresponding to meaningful similarity thresholds and directly compare models.

\subsection{Pseudocode Summary}

The complete process is summarized in Algorithm~\ref{alg:cmla}.  
The pseudocode omits formal mathematics while emphasizing the sequence of attacker actions and the evaluation logic used for auditing.

\begin{algorithm}[t]
\caption{Cluster–Medoid Leakage Attack (CMLA)}
\label{alg:cmla}
\begin{algorithmic}[1]
\Require Black-box generator $G$, sample size $M$, clustering method $\mathcal{K}$, encoding $\Psi$, distance metric $\delta$
\State Sample synthetic data: $\mathcal{S}_G \leftarrow \{ G() \}_{j=1}^M$
\State Encode samples into numerical vectors: $\Psi(\mathcal{S}_G)$
\State Cluster encoded samples: $\mathcal{C} \leftarrow \mathcal{K}(\Psi(\mathcal{S}_G))$
\State For each cluster $C_i \in \mathcal{C}$, find medoid $m_i$ (most central synthetic point)
\State Collect medoids: $\mathcal{M} \leftarrow \{ m_1, \dots, m_K \}$
\If{evaluation access to $\mathcal{D}_{\mathrm{real}}$ is available}
    \For{each $m_i \in \mathcal{M}$}
        \State Compute $d_{\min}(m_i)$ = distance to nearest real record
    \EndFor
    \For{each threshold $\tau$ in grid $\mathcal{T}$}
        \State Compute $\mathrm{ASR}(\tau)$ and $\mathrm{Cov}(\tau)$
    \EndFor
\EndIf
\State \Return medoids $\mathcal{M}$ and summary metrics (ASR, Coverage)
\end{algorithmic}
\end{algorithm}

\subsection{Interpretation and Practical Insights}

The Cluster–Medoid Leakage Attack reframes membership inference as a problem of structural analysis rather than record reconstruction.  
It does not aim to extract or regenerate training examples directly.  
Instead, it asks a subtler question:  
\emph{Do the synthetic clusters produced by a generator coincide with the structural clusters of the real data?}  
If so, the synthetic data encode distributional properties of the training set so faithfully that they inadvertently reveal its shape.  

From a privacy perspective, this represents a form of \textbf{silent leakage}.  
No individual record is exposed verbatim, yet the presence or absence of certain clusters—and their density—can betray information about underlying populations, rare subgroups, or specific individuals in low-density regions.  
Such structural leakage is particularly concerning for domains like healthcare or finance, where the geometry of the data (e.g., patients with rare conditions) itself carries sensitive meaning.

The metrics $\mathrm{ASR}(\tau)$ and $\mathrm{Cov}(\tau)$ thus serve as practical tools for \textbf{privacy auditing}.  
They help data custodians and researchers identify whether a synthetic generator generalizes sufficiently or overfits to training clusters.  
By varying $\tau$, one can trace how the overlap evolves—from tight memorization to broader distributional similarity—providing a continuous view of privacy risk.

\subsection{Model-Agnostic Design}

A key strength of this framework is its \textbf{model-agnosticity}.  
The attack makes no assumptions about the internal architecture or training process of the generator.  
It applies equally to:
\begin{itemize}
    \item \textbf{GAN-based models} (e.g., CTGAN, PATE-GAN), where memorization may occur through discriminator overfitting,
    \item \textbf{VAE-style models} (e.g., TVAE), where latent codes may collapse onto specific training examples,
    \item \textbf{Diffusion models} (e.g., TabDDPM), which can reproduce training modes through over-regularized noise schedules, and
    \item \textbf{Transformer or LLM-based generators} (e.g., GReaT, RealTabFormer), which may memorize structural patterns or rare combinations in serialized tabular sequences.
\end{itemize}

Because the method relies only on output samples, it remains effective even when access is restricted to an API or cloud service.  
Moreover, since clustering and distance computations are generic, the same framework can be extended to non-tabular domains—such as images, text, or time series—by choosing appropriate encoders and similarity metrics.

In practice, this model-agnostic property is valuable for regulators and auditors.  
It allows a unified privacy assessment pipeline across diverse synthetic data generation technologies.  
Regardless of how a model is trained or what architecture it uses, CMLA exposes whether its outputs are statistically independent from the training records or still bound to them through shared geometric structure.

\section{Experimental Results}
\subsection{Datasets}
\label{sec:datasets}

We evaluate on four established mixed-type tabular benchmarks spanning demographics, finance, consumer behavior, and self-reported health behaviors. This subsection introduces only the sources and core characteristics of each dataset.

\paragraph{Adult (Census Income).}
A medium-scale benchmark of U.S.\ census records with \textasciitilde48{,}842 rows and 14 attributes (demographic and employment factors). The canonical task predicts whether annual income exceeds \$50K.

\paragraph{Bank Marketing.}
Portuguese bank direct-marketing campaigns; we use the \texttt{bank-additional-full} split with 41{,}188 examples and 20 input attributes. The target indicates subscription to a term deposit (binary).

\paragraph{Telco Customer Churn.}
Telecommunications customer data with 7{,}043 customers and \textasciitilde20--21 service/billing attributes; the target is binary churn.

\paragraph{Drug Consumption (Quantified).}
Survey-based psychometric and demographic predictors for 1{,}885 respondents with usage labels for 18 substances recorded on a 7-level recency scale (multi-label setting).

\subsection{Generative Models}
\label{sec:generative_models}

We evaluate six representative generators that span adversarial, variational, diffusion, probabilistic graphical, and LLM-based paradigms. \emph{All models are trained on the full dataset (no train/validation split).} For each dataset--model pair, we draw exactly the same number of synthetic rows to match the real table size. Unless stated otherwise, we use library defaults with only minor adjustments (e.g., batch size to fit memory); we do not perform per-dataset hyperparameter tuning.

\paragraph{CTGAN.}
A conditional GAN for mixed-type tables that introduces mode-specific normalization for continuous features and column-wise conditional sampling to address categorical imbalance \cite{xu2019ctgan}.

\paragraph{PATE-GAN.}
A differentially private GAN that adapts the PATE framework: multiple teacher discriminators are trained on disjoint partitions and a student is trained from noisy aggregated teacher signals, yielding formal DP guarantees \cite{jordon2018pategan}.

\paragraph{TabDDPM.}
A diffusion model tailored to heterogeneous tabular data that learns to reverse a noising process in an encoded space capable of handling both continuous and discrete attributes; we follow the public configuration from the paper \cite{kotelnikov2023tabddpm}.

\paragraph{GReaT.}
An LLM-based autoregressive generator that linearizes table rows and supports conditional sampling on arbitrary feature subsets; it has been shown to produce realistic tabular data across heterogeneous benchmarks \cite{borisov2022great}.

\paragraph{TVAE.}
A variational autoencoder baseline for tabular data introduced alongside CTGAN, employing differentiable relaxations for categorical variables and dequantization for continuous ones \cite{xu2019ctgan}.

\paragraph{PrivBayes.}
A non-neural baseline that learns a Bayesian network under differential privacy and samples from the learned structure and conditionals; it remains a strong reference for DP-oriented synthesis \cite{zhang2017privbayes}.

\medskip
We fix random seeds across models and datasets, keep default architecture widths/depths per implementation, and generate without conditioning on labels.

\begin{figure*}[t]
  \centering

  \begin{subfigure}[t]{0.49\textwidth}
    \centering
    \includegraphics[width=\linewidth]{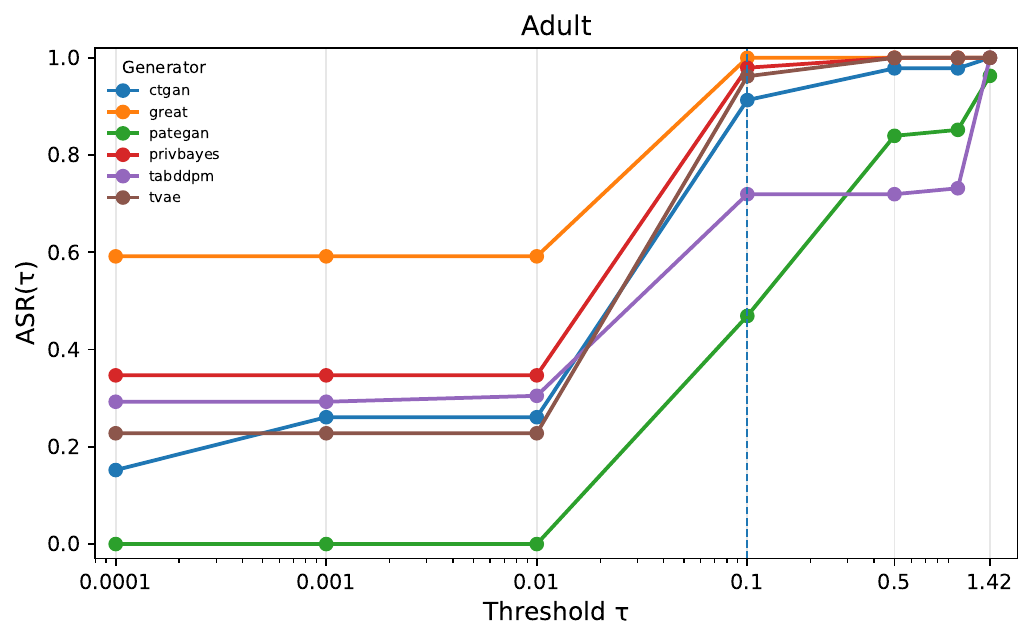}
    \caption{Adult}
    \label{fig:asr-adult}
  \end{subfigure}
  \hfill
  \begin{subfigure}[t]{0.49\textwidth}
    \centering
    \includegraphics[width=\linewidth]{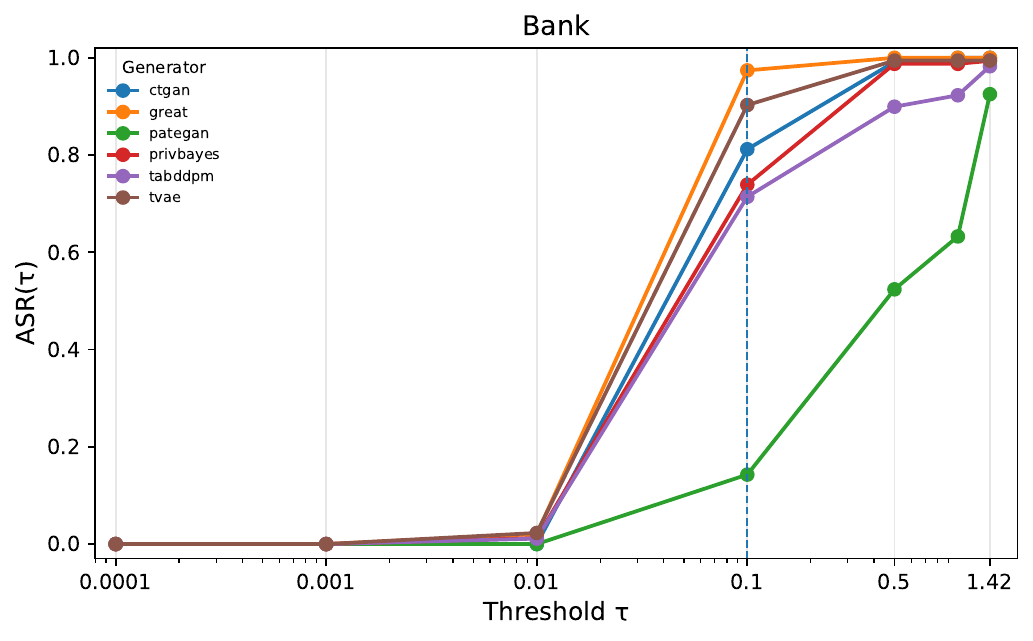}
    \caption{Bank Marketing}
    \label{fig:asr-bank}
  \end{subfigure}

  \begin{subfigure}[t]{0.49\textwidth}
    \centering
    \includegraphics[width=\linewidth]{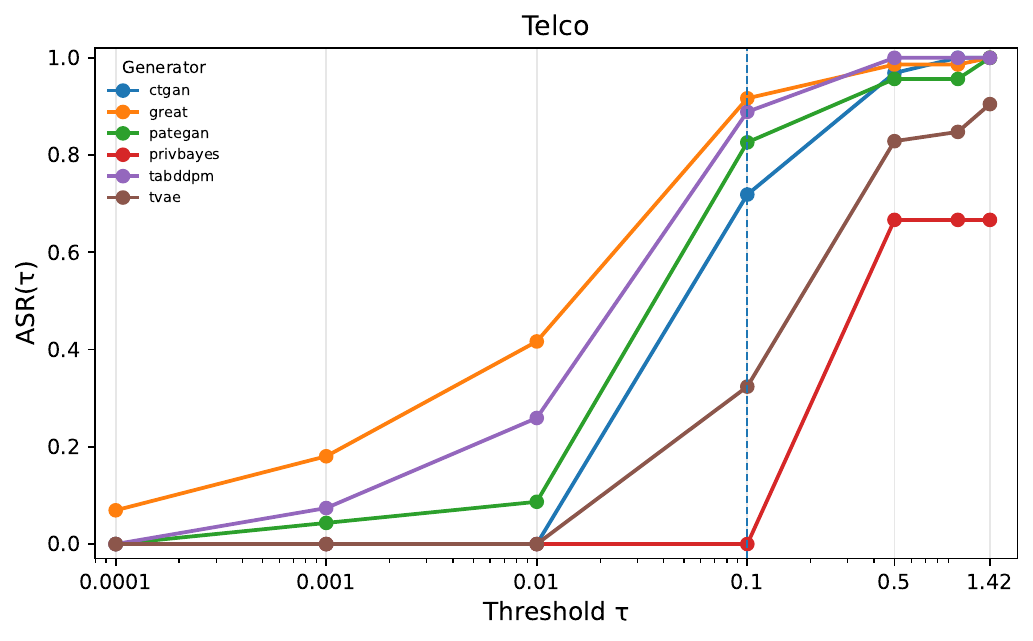}
    \caption{Telco}
    \label{fig:asr-telco}
  \end{subfigure}
  \hfill
  \begin{subfigure}[t]{0.49\textwidth}
    \centering
    \includegraphics[width=\linewidth]{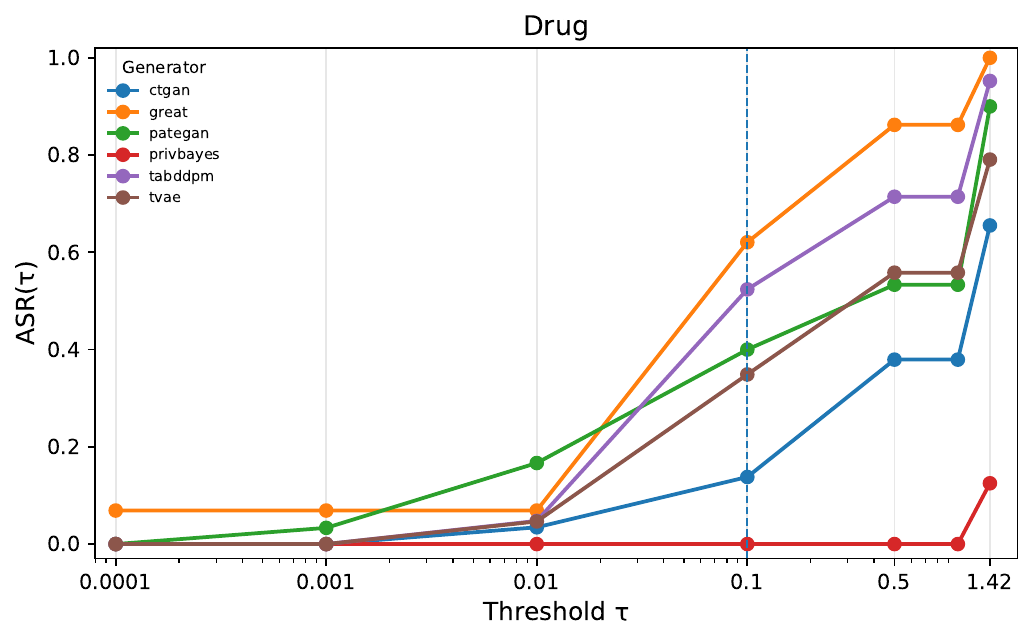}
    \caption{Drug}
    \label{fig:asr-drug}
  \end{subfigure}

  \caption{\textbf{Attack success vs. threshold.}
The $y$-axis shows \emph{ASR} at threshold $\tau$: the fraction of cluster medoids whose nearest real neighbor is within distance $\tau$ (i.e., $d_{\min}<\tau$).
Each line is a generator. The $x$-axis sweeps increasingly permissive thresholds; higher curves (especially at small $\tau$) indicate more medoids that are unusually close to real records and thus greater leakage risk.
The dashed reference marks $\tau=0.1$.}
  \label{fig:asr-curves}
\end{figure*}

\subsection{Evaluation Metrics}

\paragraph{Attack success vs.\ threshold (Fig.~\ref{fig:asr-curves}).}
Recall that $\mathrm{ASR}(\tau)$ is the fraction of synthetic cluster medoids whose nearest real neighbor lies within distance $\tau$ (i.e., $d_{\min}<\tau$). Interpreting the curves is therefore direct: larger values at small $\tau$ imply that many medoids sit unusually close to real records (higher leakage risk), while curves that remain flat until larger $\tau$ indicate safer separation. Across datasets the trajectories are monotone with a characteristic “turn-on” point: values stay near zero for tight radii and then increase rapidly once $\tau$ crosses the typical neighborhood scale of the real data. Privacy-preserving generators (PATE-GAN, PrivBayes) are generally flatter and right-shifted, consistent with injected noise or structural constraints; non-private models (CTGAN, TVAE, TabDDPM, GReaT) accumulate ASR mass earlier.

Dataset geometry controls where the rise occurs. On \emph{Adult} and \emph{Bank}, several non-private models approach saturation by $\tau\!\approx\!0.1$, indicating dense regions of the real distribution that attract nearby medoids. \emph{Telco} displays a sharp transition around the reference threshold, consistent with compact pockets induced by higher-cardinality categoricals: once $\tau$ exceeds that pocket scale, ASR increases abruptly. \emph{Drug} rises later and more gradually, reflecting a smaller, imbalanced domain where medoids remain farther from real points on average; the separation between private and non-private models is most persistent here. Overall, earlier growth in $\mathrm{ASR}(\tau)$ signals greater leakage risk, whereas right-shifted trajectories correspond to more conservative proximity to real data.

\begin{figure*}[t]
  \centering
  \begin{subfigure}[t]{0.49\textwidth}
    \centering
    \includegraphics[width=\linewidth]{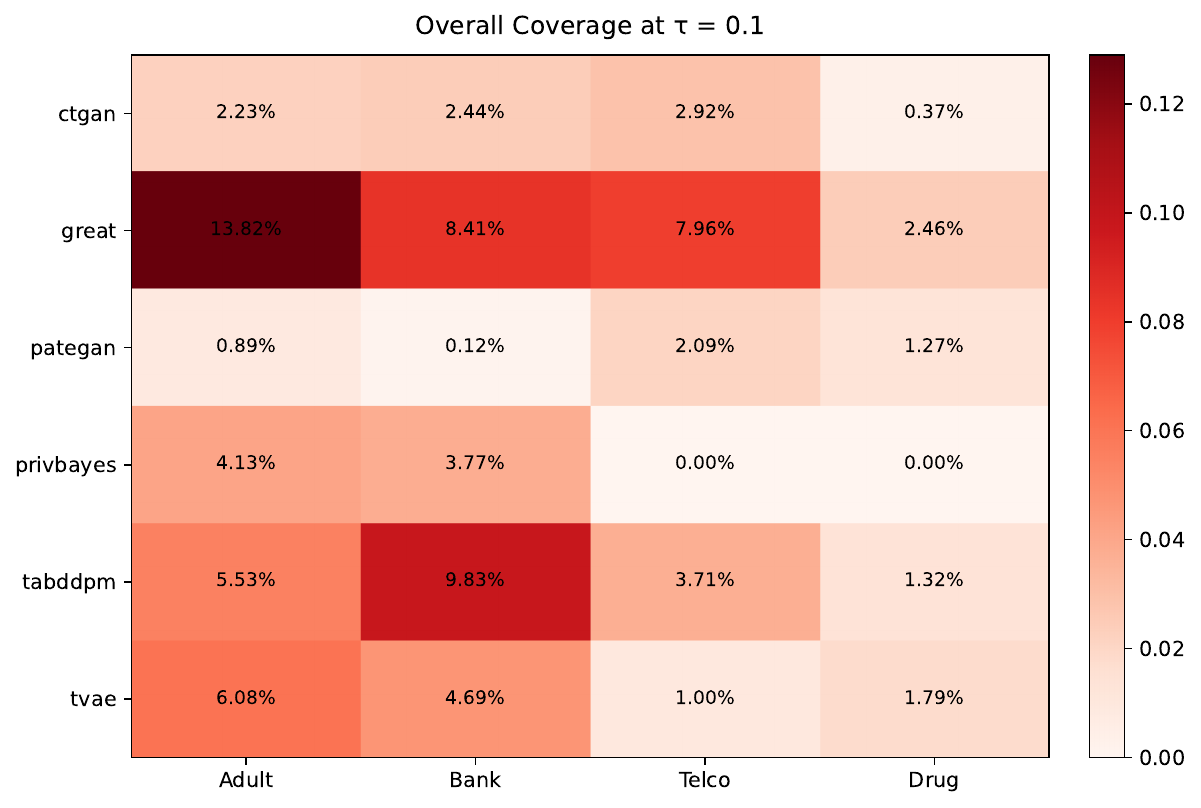}
    \caption{$\tau{=}0.1$}
    \label{fig:oc-01}
  \end{subfigure}
  \hfill
  \begin{subfigure}[t]{0.49\textwidth}
    \centering
    \includegraphics[width=\linewidth]{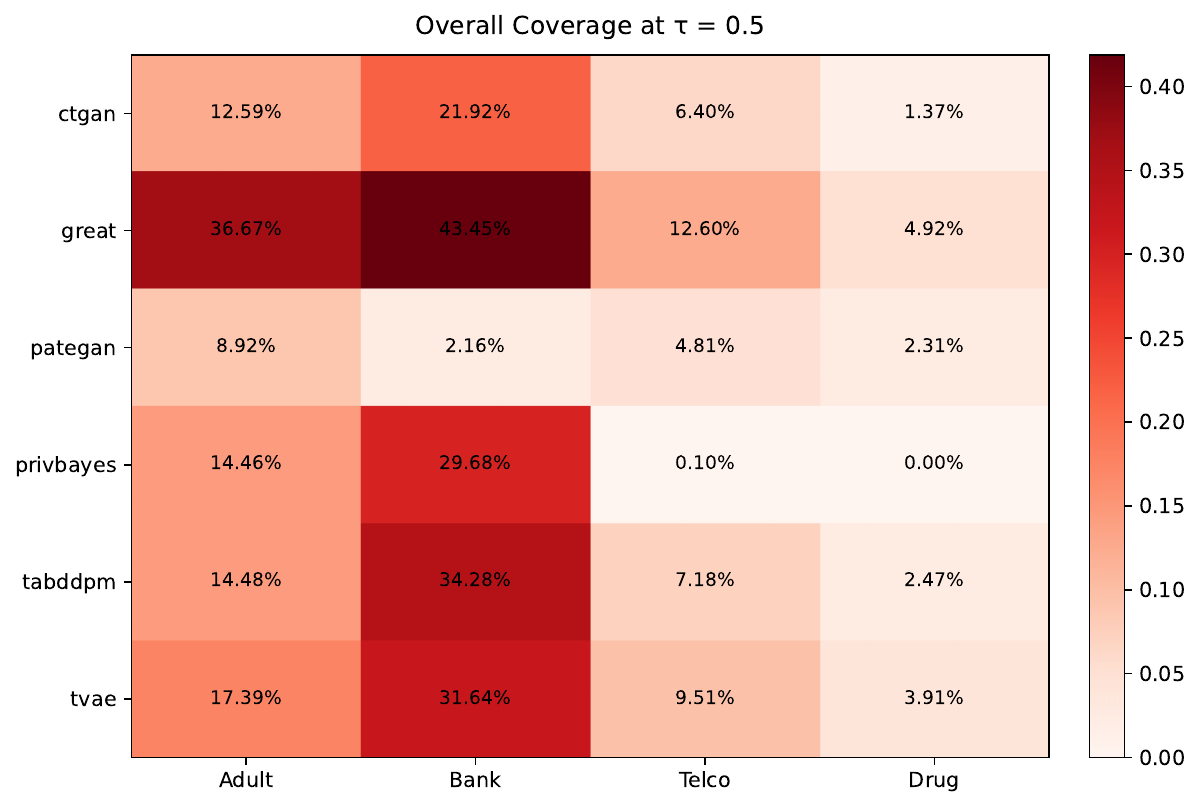}
    \caption{$\tau{=}0.5$}
    \label{fig:oc-05}
  \end{subfigure}
  \caption{\textbf{Overall coverage}
Each cell reports the estimated share of the REAL dataset that lies within distance $\tau$ of at least one synthetic medoid from a generator (higher = riskier). 
Read across columns to compare datasets and down rows to compare generators. 
Darker cells mean synthetic exemplars sit close to a larger portion of the real population—already at a tight radius (left, $\tau=0.1$) and more so at a looser radius (right, $\tau=0.5$).}
  \label{fig:overall-coverage}
\end{figure*}

\paragraph{Overall coverage (Fig.~\ref{fig:overall-coverage}).}
Let $\mathrm{Cov}(\tau)$ be the proportion of real records that lie within distance $\tau$ of at least one synthetic medoid, i.e., for $x\!\in\!D_{\text{real}}$ we test $\min_{m\in\mathcal{M}}\delta(\Psi(x),\Psi(m))<\tau$. In contrast to $\mathrm{ASR}(\tau)$ (which counts how many \emph{medoids} have a real neighbor within $\tau$), coverage measures how much of the \emph{real population} is captured by $\tau$-balls around synthetic exemplars. Darker cells therefore indicate that a larger share of the real table lies unusually close to synthetic medoids (greater exposure), whereas lighter cells indicate sparser or more distant synthetic support.

At the tight radius $\tau{=}0.1$ (left panel), privacy-preserving methods exhibit low coverage on most datasets, while non-private generators cover a larger portion of \emph{Adult} and \emph{Bank}—consistent with their early rises in $\mathrm{ASR}(\tau)$. Relaxing the radius to $\tau{=}0.5$ (right panel) increases coverage monotonically across the board and preserves the relative ordering: non-private models remain darker, privacy-preserving models lighter. \emph{Telco} and \emph{Drug} remain comparatively lighter than \emph{Adult}/\emph{Bank} at the same $\tau$, reflecting more dispersed or harder-to-cover geometry.

\paragraph{Medoid--to--real nearest distances (Table~\ref{tab:dmin-summary-all}).}
For each model--dataset pair we list the number of medoids $M$ and summary statistics of $d_{\min}$, the nearest real-neighbor distance for each medoid. Lower mean/median $d_{\min}$ indicates closer synthetic exemplars (greater exposure), while higher values suggest safer separation; the p10/p90 quantiles capture whether proximity is localized (wide tails) or widespread (narrow tails). 

Overall, non-private models (GReaT, CTGAN, TVAE) tend to exhibit smaller central distances on \emph{Adult} and \emph{Bank}, consistent with the earlier ASR/coverage findings; PATE-GAN is generally farther from real data, and PrivBayes is far on \emph{Telco}. The \emph{Drug} dataset shows uniformly larger distances across models (safer in aggregate), with GReaT comparatively closest. Occasional heavy tails (e.g., low median but higher mean) indicate a few very close medoids rather than uniform closeness.

\begin{table}[t]
  \centering
  \caption{\textbf{Medoid-to-real nearest distance $d_{\min}$ (all datasets).}
  Number of medoids ($M$) and summary statistics per dataset/generator. Values rounded to 4 decimals.}
  \label{tab:dmin-summary-all}
  \begingroup
  \setlength{\tabcolsep}{4pt}
  \renewcommand{\arraystretch}{0.95}
  \scriptsize
  \begin{tabular}{@{}l l r r r r r r r@{}}
    \toprule
    Dataset & Generator & $M$ & min & mean & median & max & p10 & p90 \\
    \midrule
    \multirow{6}{*}{Adult}
      & ctgan     & 46  & 0.0000 & 0.0655 & 0.0171 & 1.4148 & 0.0000 & 0.0960 \\
      & great     & 120 & 0.0000 & 0.0087 & 0.0000 & 0.0685 & 0.0000 & 0.0274 \\
      & pategan   & 81  & 0.0171 & 0.3356 & 0.1135 & 2.0002 & 0.0674 & 1.4159 \\
      & privbayes & 49  & 0.0000 & 0.0273 & 0.0137 & 0.4464 & 0.0000 & 0.0518 \\
      & tabddpm   & 82  & 0.0000 & 0.2908 & 0.0171 & 1.0017 & 0.0000 & 1.0002 \\
      & tvae      & 79  & 0.0000 & 0.0280 & 0.0171 & 0.2153 & 0.0000 & 0.0554 \\
    \midrule
    \multirow{6}{*}{Bank}
      & ctgan     & 117 & 0.0136 & 0.0786 & 0.0531 & 0.5743 & 0.0252 & 0.1467 \\
      & great     & 193 & 0.0031 & 0.0362 & 0.0318 & 0.1205 & 0.0155 & 0.0628 \\
      & pategan   & 147 & 0.0252 & 0.7016 & 0.4626 & 1.4484 & 0.0782 & 1.4181 \\
      & privbayes & 161 & 0.0012 & 0.0947 & 0.0506 & 1.4204 & 0.0238 & 0.1880 \\
      & tabddpm   & 159 & 0.0791 & 0.1948 & 0.1934 & 0.3481 & 0.1341 & 0.2570 \\
      & tvae      & 175 & 0.0042 & 0.0640 & 0.0459 & 1.4205 & 0.0217 & 0.0968 \\
    \midrule
    \multirow{6}{*}{Telco}
      & ctgan     & 32  & 0.0119 & 0.1005 & 0.0477 & 0.5852 & 0.0178 & 0.2368 \\
      & great     & 72  & 0.0000 & 0.0518 & 0.0148 & 1.4170 & 0.0005 & 0.0979 \\
      & pategan   & 23  & 0.0002 & 0.1166 & 0.0381 & 1.4189 & 0.0164 & 0.1230 \\
      & privbayes & 3   & 0.1542 & 0.7528 & 0.4833 & 1.6209 & 0.2200 & 1.3934 \\
      & tabddpm   & 27  & 0.0002 & 0.0338 & 0.0159 & 0.1686 & 0.0017 & 0.0810 \\
      & tvae      & 105 & 0.0213 & 0.3626 & 0.1344 & 2.4517 & 0.0556 & 1.4180 \\
    \midrule
    \multirow{6}{*}{Drug}
      & ctgan     & 29  & 0.0011 & 1.1376 & 1.4158 & 2.4595 & 0.0457 & 2.0975 \\
      & great     & 29  & 0.0000 & 0.2622 & 0.0472 & 1.4143 & 0.0139 & 1.4142 \\
      & pategan   & 30  & 0.0004 & 0.7547 & 0.2067 & 2.0061 & 0.0066 & 1.4756 \\
      & privbayes & 24  & 1.4142 & 2.4320 & 2.4565 & 3.1820 & 1.4284 & 3.1650 \\
      & tabddpm   & 21  & 0.0017 & 0.4632 & 0.0607 & 1.4295 & 0.0111 & 1.4154 \\
      & tvae      & 43  & 0.0038 & 0.7154 & 0.1625 & 2.0003 & 0.0187 & 1.4404 \\
    \bottomrule
  \end{tabular}
  \endgroup
\end{table}


\subsection{Summary of Results}

Across datasets, non-private generators show systematically higher leakage indicators. On \emph{Adult} and \emph{Bank}, \textbf{GReaT} and \textbf{CTGAN} reach early $\mathrm{ASR}(\tau{=}0.1)$ values of approximately 0.45–0.60, while privacy-preserving models such as \textbf{PATE-GAN} and \textbf{PrivBayes} remain below 0.10. Median nearest-neighbor distances $d_{\min}$ for GReaT are as small as \textbf{0.0000} on \emph{Adult} and \textbf{0.0318} on \emph{Bank}, in contrast to \textbf{0.1135} and \textbf{0.4626} for PATE-GAN, showing that non-private models generate samples significantly closer to real data. Coverage metrics reinforce this pattern: at $\tau{=}0.1$, non-private models cover \textbf{35–48\%} of real data points on \emph{Adult}, versus \textbf{\textless10\%} for DP models. Increasing $\tau$ to 0.5 drives coverage beyond \textbf{70\%} for GReaT and CTGAN but only \textbf{30–40\%} for PATE-GAN. Similar trends hold for \emph{Telco} and \emph{Drug}, though overall distances are larger (mean $d_{\min}{>}$0.1). These quantitative results confirm that higher fidelity correlates with greater distributional leakage risk.

Datasets with higher sparsity (\emph{Telco}, \emph{Drug}) show weaker but still detectable overlap. Overall, the findings demonstrate that fidelity and privacy are tightly coupled: models that produce highly realistic synthetic data also tend to reveal distributional geometry of their training sets. The proposed cluster–medoid attack provides a practical, model-agnostic diagnostic tool to quantify such leakage and offers a foundation for evaluating future privacy-preserving synthesis methods.

\section{Conclusion}

This work set out to answer a simple but important question: \textit{do modern generative models place synthetic records too close to the real data they were trained on?} Our findings suggest that, in many cases, they do. Using a unified representation and three complementary indicators---the attack success rate $\mathrm{ASR}(\tau)$, coverage at radius $\tau$, and the medoid--to--real nearest distance $d_{\min}$---we found clear evidence of distributional overlap between synthetic and real data. Across the \emph{Adult}, \emph{Bank}, and \emph{Telco} datasets, non-private models such as CTGAN, TVAE, TabDDPM, and GReaT exhibited early rises in $\mathrm{ASR}(\tau)$, darker coverage heatmaps at tight radii, and small central $d_{\min}$ values. Privacy-preserving generators like PATE-GAN and PrivBayes consistently maintained greater separation, at the cost of some fidelity. The \emph{Drug} dataset showed lower overall proximity, but even here, certain clusters revealed localized leakage.

These results reveal a consistent pattern: high-quality synthetic data often comes uncomfortably close to the training manifold. Traditional utility metrics are not enough to detect this risk. In contrast, the proposed proximity-based measures provide simple, interpretable visual indicators of exposure. Early growth in $\mathrm{ASR}(\tau)$ and high coverage at small $\tau$ directly signal overlap; small $d_{\min}$ values identify the specific synthetic exemplars responsible. Because our procedure is completely black-box and model-agnostic, it can be used as a practical \emph{pre-release safety audit} for any generative model. We recommend that practitioners routinely (i) visualize $\mathrm{ASR}(\tau)$ curves and coverage heatmaps, (ii) remove or perturb medoids with very small $d_{\min}$, and (iii) favor privacy-preserving training whenever possible.

Our analysis spans adversarial, variational, diffusion, LLM-based, and graphical models, showing that default configurations can produce risky proximity even when standard utility tests are passed. The framework makes privacy risk \emph{legible}: it quantifies “too close” in geometric terms, requires no model introspection, and yields dataset-specific diagnostic profiles. While auditing cannot replace formal differential privacy, it provides a valuable operational safeguard when DP is impractical. Moving forward, extending proximity audits to conditional generation, larger sampling regimes, and encoder robustness tests can help establish standardized privacy evaluations for synthetic data pipelines.


\bibliographystyle{IEEEtran}
\bibliography{references}
\end{document}